%% file: main.tex
\documentclass{article} 
\usepackage{iclr2021_conference,times}

\input{math_commands.tex}

\usepackage{hyperref}
\usepackage{url}
\usepackage{amssymb}
\usepackage{graphicx}
\usepackage{authblk}

\graphicspath{ {./old/} }

\title{Syft 0.5: A Platform for Universally Deployable Structured Transparency}

\author[3,2]{Adam James Hall}
\author[2]{Madhava Jay}
\author[2]{Tudor Cebere}
\author[2]{Bogdan Cebere}
\author[2,4]{Koen Lennart van der Veen}
\author[2,5]{George Muraru}
\author[2,6]{Tongye Xu}
\author[2]{Patrick Cason}
\author[3,2]{William Abramson}
\author[2,7]{Ayoub Benaissa}
\author[2]{Chnimay Shah}
\author[2]{Alan Aboudib}
\author[2,8,9,10,11,12]{Théo Ryffel}
\author[2]{Kritika Prakash}
\author[2,12]{Tom Titcombe}
\author[2]{Varun Kumar Khare}
\author[2]{Maddie Shang}
\author[2]{Ionesio Junior}
\author[2]{Animesh Gupta}
\author[2]{Jason Paumier}
\author[2]{Nahua Kang}
\author[2]{Vova Manannikov}
\author[1,2]{Andrew Trask}

\affil[1]{University of Oxford}
\affil[2]{OpenMined}
\affil[3]{Edinburgh Napier University}
\affil[4]{Memri}
\affil[5]{Politehnica University of Bucharest}
\affil[6]{Philisense}
\affil[7]{École Supérieure en Informatique, Sidi Bel Abbès}
\affil[8]{INRIA Paris}
\affil[9]{ENS}
\affil[10]{PSL University}
\affil[11]{Arkhn}
\affil[12]{Tessella}

\iclrfinalcopy 
\begin{document}

\maketitle

\begin{abstract}

We present Syft 0.5, a general-purpose framework that combines a core group of privacy-enhancing technologies that facilitate a universal set of structured transparency systems. This framework is demonstrated through the design and implementation of a novel privacy-preserving inference information flow where we pass homomorphically encrypted activation signals through a split neural network for inference. We show that splitting the model further up the computation chain significantly reduces the computation time of inference and the payload size of activation signals at the cost of model secrecy. We evaluate our proposed flow with respect to its provision of the core structural transparency principles. 
\end{abstract}

\section{Introduction}

The centralisation of resources in machine learning information flows forces a Pareto efficiency trade-off between data utility and privacy. When participating in these flows, data owners cannot know that their data has not been sold, retained for far longer than intended or otherwise used for purposes outside the understood context-relative informational norms \cite{nissenbaum2004privacy}. A key risk factor here is the data copy problem. Data once copied, even by well meaning actors, cannot ever be guaranteed to have been destroyed, as has become painfully clear by the take down of the DukeMTMC dataset (\cite{Ristani2016PerformanceMA}). Controversy surrounding the unintended usage of the public dataset was raised when faces of ordinary people were used without permission to serve questionable ends. This resulted in the data being revoked, breaking the social context of the data subject's participation in that flow (\cite{princeton2020cfitp}). Copies of the data and derivative datasets were still freely available on the internet and academic papers are still being published. This is such a pervasive issue that websites exist for tracking the leakage and misuse of peoples facial data for inappropriate, extra-contextual use cases (\cite{Exposing.ai}).

The increasing prevalence of organised criminal groups online remains a risk, forcing centralised data processors to respect informational norms and securely store the user's data. Adequate maintenance of confidentiality, integrity, and availability of data represents an increasing liability for processors. Between 2015 and 2020, cybersecurity spending worldwide almost doubled from \$75.6 billion to \$124 (\cite{lee2021cybersecurity}). The recent General Data Protection Regulation (GDPR) legislation (\cite{EU}) requires explicit consent from data subjects to allow the processing of their data for any purpose. GDPR and similar laws represent a formidable bureaucratic overhead to researchers who process data concerning EU citizens. This issue is further compounded by a lack of trust between data subjects and data processors. Under these circumstances, research on private data is blocked due to privacy ramifications- without ever contributing to an information flow's contextually driven purpose or values. If these obstacles are removed, research goes ahead with potentially disastrous social and political consequences for the data subjects and their community.

In this work, we contribute an open-source, universally deployable framework capable of facilitating information flows that cater to the core tenets of Structured Tranansparency (ST); giving actors the confidence to participate with sensitive information. Our tool provides a broader gamut of privacy-enhancing technologies (PETs) than existing frameworks (Table \ref{table:feature_comparison}) with the ability to accommodate ST in an increasingly nuanced set of information flows. We define a novel, structurally transparent information flow that utilises the split learning technique to reduce computation overhead in homomorphically encrypted inference.

\section{Framework Description}

\begin{figure}[h]
\begin{center}
\fbox{\rule[0cm]{0cm}{0cm} \includegraphics[width=13.5cm]{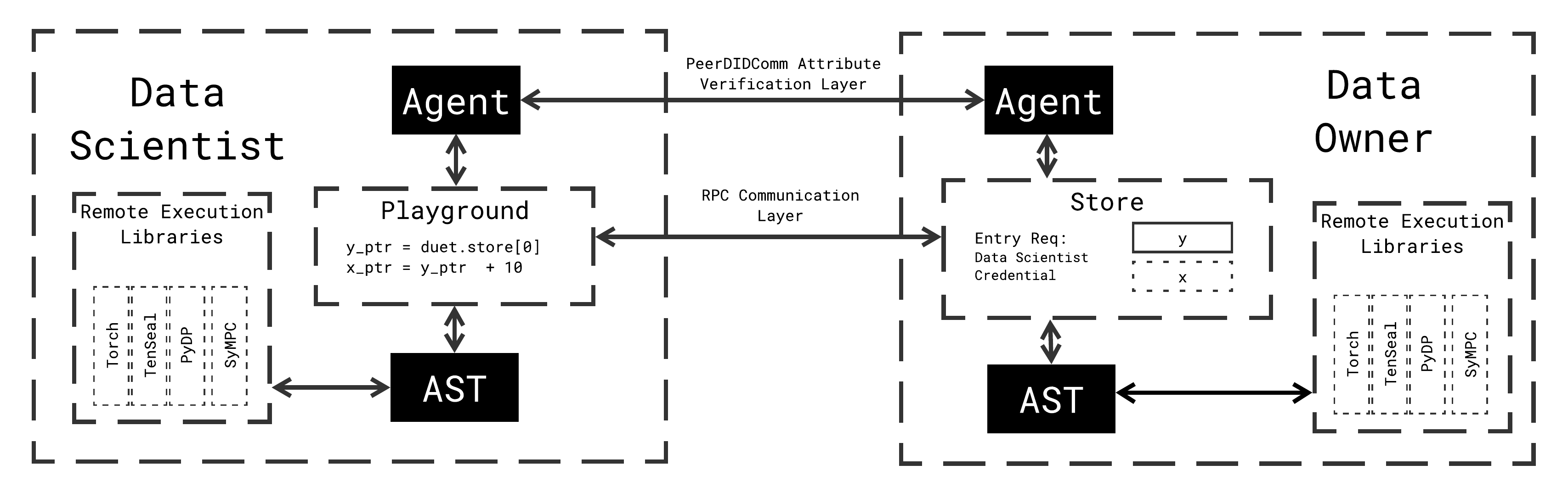} \rule[0cm]{0cm}{0cm}}
\end{center}
\caption{Duet Architecture Diagram}
\label{figure:architecture}
\end{figure}

\cite{}

Syft allows data scientists to perform compute data they don't own and cannot see. This is achieved through the separation of concerns between where data resides and where code is executed. Syft allows for remote computation on any object or library that has been mapped to the Abstract Syntax Tree (AST). From a user perspective, remote variables are operated upon locally as network pointers. Operations executed on these network pointers are encapsulated as a remote procedure call performed on the remote machine. Agents can be referenced to resolve the verifiable attributes of credentials or data and models in the store without exposing a persistent identity- facilitating zero-knowledge access management as outlined in \cite{Abramson_2020}. 


After a WebRTC connection is established between peers, privacy is guaranteed through the remote execution model in which you compute operations on data without revealing it or the underlying permission scheme. Even if data is shared, it can still be encrypted using our secure multi-party computation solution, SyMPC, or homomorphic encryption through TenSEAL. Intermediary computation and results are held in a store controlled by the data owner. The data owner is able to approve or deny requests to access the data in the store by either manual approval (by hand) or automatic approval via a policy.

As an API design, Syft aims to provide a near transparent layer around supported frameworks and libraries so that data scientists don't need to learn new tools to develop their algorithms in a privacy-friendly fashion. Users should expect their existing code to run with minimal change and to find writing new code within Syft intuitive. Syft significantly lowers the entry-level of access to cryptography, differential-privacy and distributed computing for researchers and practitioners.

\section{Structured Transparency}

ST is an extension of the contextual integrity framework (\cite{nissenbaum2009privacy}). Contextual integrity views privacy in terms of information flowing between actors in roles governed by context-relative informational norms. Nissembaum's central thesis is that people feel their privacy is violated when these informational norms are broken, for example when unanticipated information is learnt by unexpected actors (\cite{nissenbaum2009privacy}). The framework leaves room for positive changes to these informational norms, suggesting that if entrenched contextual integrity is breached by new systems then they should be evaluated against whether these changes contribute to the purpose and values of the context. Contextual integrity emphasises the importance of privacy, while challenging older definitions of privacy as either control or access to information \cite{gavison1980privacy,westin1968privacy}. Instead, privacy is presented as the appropriate flow of information relative to a context, hence including control and access within a broader social framing (\cite{nissenbaum2004privacy,nissenbaum2009privacy}).

ST  extends contextual integrity's definition of an information flow by introducing a more nuanced set of properties of the flow. These properties take into account new possibilities for structuring the context and relative informational norms around an information flow. PETs implemented in Syft and presented in this paper, explicitly seek to facilitate these information flows. Information flows from a sender to a recipient and the information pertains to a data subject, who could possibly not be the sender. The context within which this information flow is embedded prescribes the roles, activities, norms and purposes for which information flows take meaning from and are justified against.  ST proposes additional definitions such as for input and output privacy, input and output verification and flow governance (\cite{TraskandEmma}). 

We seek to structure the transparency of an information flow by combining novel PETs for computation \cite{cramer2015secure}, encryption (\cite{gentry2009fully,cheon2017homomorphic}) and authentication (\cite{diffie1976new,camenisch2001efficient}). Such techniques can give confidence to the sender, recipients and subjects of an information flow that the information is integrity assured, verifiably from a sender and only visible to the set of actors and roles defined in the flow structure. Specifically, concerning machine learning use cases, the information contributed to a flow shouldn't be revealed or leaked at any step, only derivatives of this information aggregated to extract features without compromising the underlying information.

\section{Encrypted Split-Inference with Structural Transparency Guarantees}

Our work is demonstrated through Duet, a 2-party scenario using the Syft framework. One party with ownership over data provides limited access to a model owned by another actor in order to conduct inference. We define an information flow wherein a data subject receives the result of a prediction performed on their private data using another entity's private model. These private information assets owned by the Data Owner (DO) and Data Scientist (DS) respectively must interact to produce a prediction to be consumed by the DO. In a societal context, this may be governed under the same norms as a user consuming inference as a service. We use this context to evaluate the ST guarantees of the encrypted inference flow in Figure \ref{figure:information_flow}. In the actual experiment we use MNIST (\cite{MNIST}) images as data and a deep convolutional neural network as our model.

\begin{figure}[h]
\begin{center}
\fbox{\rule[0cm]{0cm}{0cm} \includegraphics[width=13.5cm]{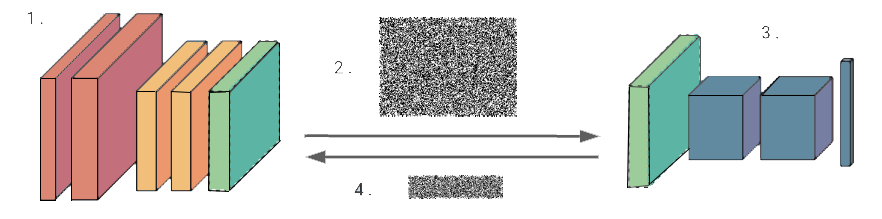} \rule[0cm]{0cm}{0cm}}
\end{center}
\caption{Private Inference Flow showing enccrypted activation signals being sent away and processed by a remote server. 1. DO segment, 2. Encrypted Activation Signal 3. DS segment 4. Encrypted Output}
\label{figure:encrypted_split_inference}
\end{figure}

\textbf{Governance -} Before this information flow may begin, we establish the context between actors and the norms that flow from this context. This is performed in step 1 of Figure \ref{figure:information_flow} where DO and DS verify attributes about one another. In Appendix A.1, Figure \ref{figure:aries_flow}, the DS proves they are a DS using the verifiable credential they received from an authority. Similarly, the DO could declare their role as a DO suitable to participate. The social context of this flow is defined through the presentation of their relevant attributes under CL Signatures (\cite{camenisch2001efficient}) through their Agent.

\textbf{Input Verification - } This section refers to steps two and three in Figure \ref{figure:information_flow}, where DO and DS load their model and data. The parameters are composed of a model and some data; both of which are private. In this case, the DO is also the consumer of the inference so we are not concerned with their inputs in our threat model. We do however, want to make sure that the DS has the model we're interested in. This is possible through storing the model object in an Agent like a verifiable credential. Similarly, this can be done with datasets to define verifiable properties like completeness, provenance and scheme conformity.

\textbf{Input Privacy - }This section refers to steps four to eight in Figure \ref{figure:information_flow} and is described visually in Figure \ref{figure:encrypted_split_inference} which describes the private inference flow. In order to maintain the privacy of the DO's data and the DS's model, the DO could encrypt their data and send it to the DS for private inference using CKKS Homomorphic Encrpytion (\cite{cheon2020remark, cheon2017homomorphic}) \cite{DBLP:journals/corr/abs-1911-11377}. However, high dimensional data incurs a significant computation overhead during encrypted computation and increased computation depth necessitates a larger polynomial modulus- increasing ciphertext size and computation requirements. Alternatively the DS may opt to only share a portion of their model with DO and execute a split neural network (SplitNN) training flow \cite{vepakomma2018split,vepakomma2018peek}. Data remains with the DO at inference time and only activation signal is shared. However, statistical information still persists in activation signals, despite being representing less information than the original data. The information contained in activation signals can be used by a malicious DS to exploit information leakage \cite{vepakomma2019reducing}. The Shredder technique (\cite{mireshghallah2020shredder}) may also be used to apply noise to activation signals before transit, however, this does not provide as strong input privacy as encryption. 

A hybrid approach is proposed which offers a trade-off between computational complexity and model privacy. We allow the DO to compute a portion of inference in plaintext and a the latter portion as ciphertext with the DS. The more model layers that are shared to the DO, the less computation depth is needed and the modulus coefficient is minimised. This results in transmitted ciphertext size and computation times which are dramatically reduced from 269.86KB to 139.6KB and 4.17s to 97ms respectively (Appendix Tables \ref{table:experiment1_results} and \ref{table:experiment2_results}). However, sharing too many layers exposes the model to theft.

\textbf{Output Privacy -} is achieved here with respect to the output recipient in step 9 of Figure \ref{figure:information_flow} where the DO decrypts the inference result locally. However, the real issue here is not that the output of the flow may reveal the DO's inputs, the DO is the consumer of the output. It's that over enough inferences, the DO may be able to infer membership of elements in the training set or perform model inversion attacks on the model (\cite{shokri2017membership, chen2020improved, Zhang_2020_CVPR}). To protect the privacy of the information in the model, we train using the Opacus library and the PrivacyEngine utility it provides for differentially private stochastic gradient descent (\cite{Opacus}). This allows us to fit our model to the general curve of the dataset rather than over-fitting. This obstructs an attackers ability to leverage the data in that data model (\cite{abadi2016deep}). This DP training step is considered a step zero and is not included in this inference flow. However, it is implemented in our experiment source code. 

\textbf{Output Verification -} In this flow output verification may relate to the removal of any bias in the model or data. The DO is the only stakeholder in the output and we can trust his data, however we should be concerned with veracity of the DS model. At the moment, we rely on the credibility of the authority that attested DS credential in step 1 of Figure \ref{figure:information_flow} where DO and DS exchange credentials. This isn't fine-grained trust. However, with Aries infrastructure we can define and deploy schemes which may track the varacity and performance of models for presentation to inference consumers fit to the requirements of any regulation for model governance use cases. 

\section{Conclusions}
Contextual integrity and ST of information flows provide the theoretical framework for privacy at scale. While privacy has traditionally been viewed as a permission system problem alone, Syft delivers privacy at all levels; by leveraging multiple cryptographic protocols and distributed algorithms, it enables new ways to structure information flows. The work being demonstrated is one such novel flow, containing ST guarantees in a 2-party setting, or Duet. Duet provides a research-friendly API for a Data Owner to privately expose their data, while a Data Scientist can access or manipulate the data on the owner's side through a zero-knowledge access control mechanism. This framework is designed to lower the barrier between research and privacy-preserving mechanisms, so that scientific progress can be made on data that is currently inaccessible or tightly controlled. In future work, Syft will integrate further with PyGrid (\cite{PyGrid}), to provide a user interface for organisational policy controls, cloud orchestration, dashboards and dataset management; to allow research institutions to utilise Syft within their existing research data policy frameworks.



\bibliography{iclr2021_conference}
\bibliographystyle{iclr2021_conference}

\section{Acknowledgments}
The authors would like to thank the OpenMined community, without which PySyft would not be possible. In particular, the authors of this paper would like to give acknowledge to contribution of; Karl Higley, Anshuman Singh, Anubhav Raj Singh, Ariann Farias, Arturo Marquez Flores, Avinash Swaminathan, Ben Fielding, Chirag Gomber, Chitresh Goel, Harkirat Singh, Hideaki Takahashi, Irina Bejan, Jason Paulmier, Jirka Borovec, Joel Lee, Lee Yi Jie Joel, Nahua Kang, Nicolas Remerscheid, Param Mirani, Plamen Hristov, Raghav Prabhakar, Rima Al Shikh, Vaibhav Vardhan, Wansoo Kim and Zarreen Naowal Reza. 

\appendix

\section{Appendix}

\subsection{Aries Agent Verification flow}

\begin{figure}[h]
\begin{center}
\fbox{\rule[0cm]{0cm}{0cm} \includegraphics[width=13.5cm]{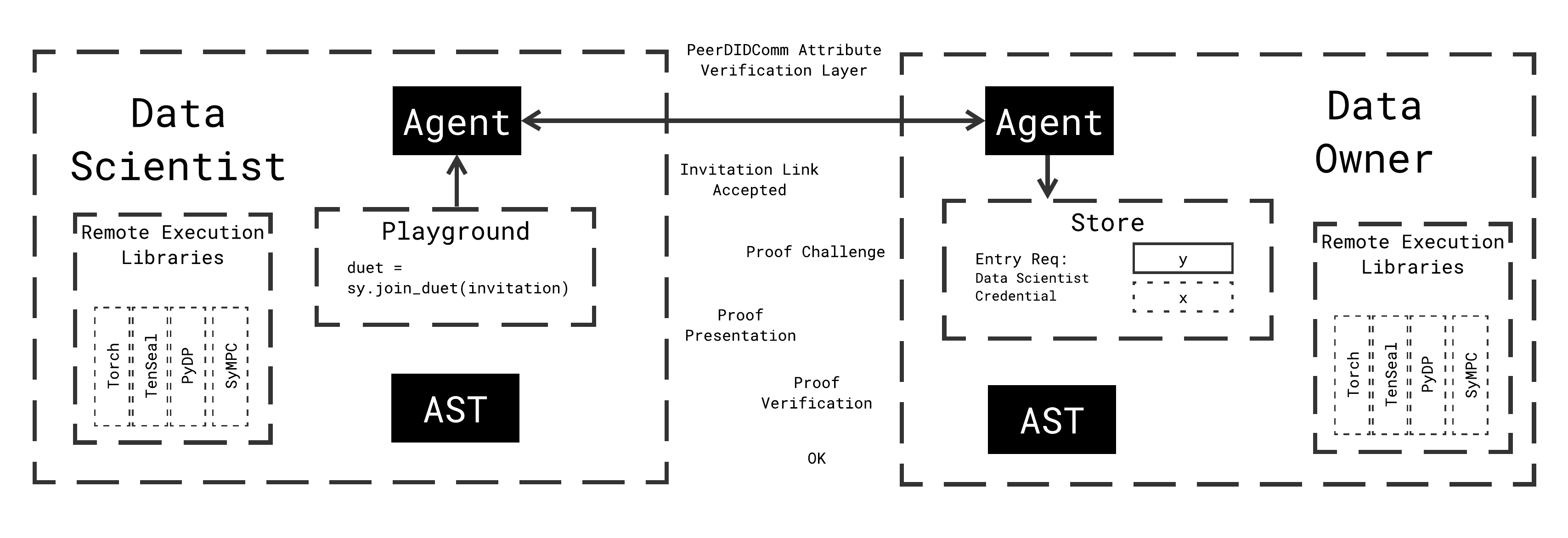} \rule[0cm]{0cm}{0cm}}
\end{center}
\caption{Aries Agent Verification flow}
\label{figure:aries_flow}
\end{figure}

\subsection{Information flow}

\begin{figure}[h]
\begin{center}
\fbox{\rule[0cm]{0cm}{0cm} \includegraphics[width=13.5cm]{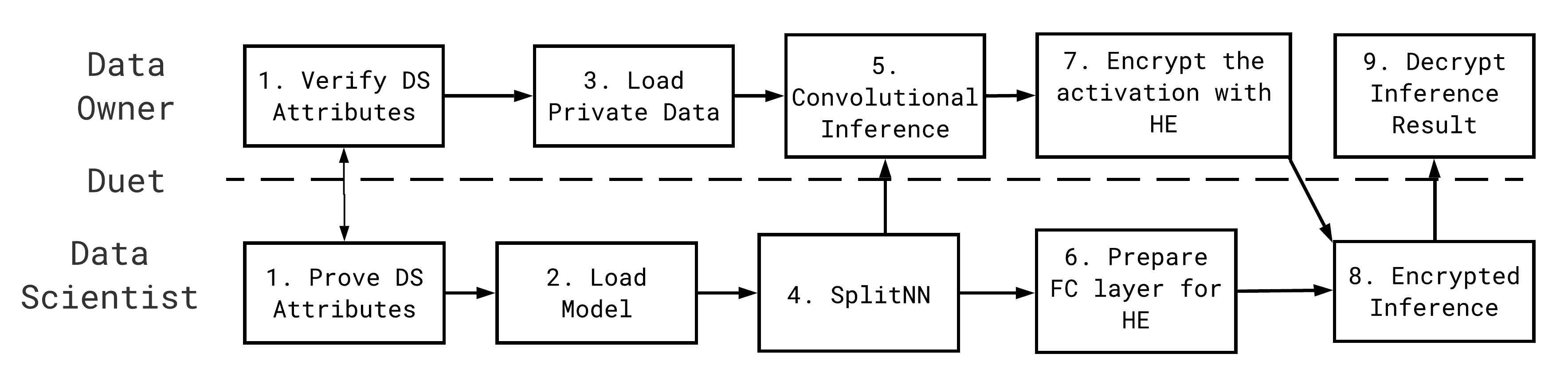} \rule[0cm]{0cm}{0cm}}
\end{center}
\caption{Private Inference Flow}
\label{figure:information_flow}
\end{figure}

\subsection{Architecture Concepts}

\textbf{Data Owner / Data Scientist - }
The Data Owner (DO) is the Duet session creator, the party that has implicit control over the store and permissions. The Data Scientist (DS), is the general terminology for the peer node which connects to a DO, and has tightly controlled limitations on access to objects in the store. There are multi-party configurations of peers which provide more complex scenarios which we won't detail here.

\textbf{Node \& Client - }
In Syft the smallest unit of secure communication and computation is a Node. These Nodes are capable of running a variety of services. By registering new message types and code that executes on receipt of those messages; services augment the capability of a Node with additional functionality. To ensure validity, all messages are signed by their sender and include a verification key. Signing is done with a PyNaCl a Python libsodium wrapper using 256-bit Ed25519. When a message arrives it is verified and then the delivery address is checked. Nodes are designed to be aware of other Nodes, and are able to forward on messages they receive to other known addresses. Once a message arrives at its destination, it is delivered to the appropriate service for execution.

Nodes have the concept of a root verification key and all services can opt to allow execution by messages signed by root clients only. In the event of an error, exceptions can be transmitted back to the sender, defaulting to UnknownPrivateException and allowing for careful exposure of real exceptions when safe and relevant. Nodes can optionally contain a key-value store, which can be backed with memory or disk. 

To provide a public API interface for users to communicate with a Node, we have a Client which provides a handle to easily invoke remote functions and view metadata about a remote Node. While there are two Nodes in the Figure 1 Duet Architecture, there is asymmetry with the DO side having a Store and both Clients having a handle to the DO Node. In this sense all transfer of data is handled by requests to the DO node and approvals by the DO root client to those requests. Additionally the DO's Client never explicitly sends data or execution requests to the DS Node. This configuration provides a streamlined work flow in scenarios where one side has private data they wish to host.

While this is one possible topology, due to the flexible design, other architectures of two or more parties are possible over any size or shape of network. By simply adding additional services to a Node, entirely new networked functionality can be constructed. This mechanism is how PyGrid is able to augment Syft to create Cloud and on-prem worker distribution and data storage. 

\textbf{The Store - }
Duet shares data through the serialization of objects as Protobuf messages. These messages are wrapped in a Storable Object (SO) interface that handles metadata and interaction with a Store. The Store may be located either in-memory or on disk. In Duet, the Store is located within the Python process of the DO. However, because all the store interaction is done through serialized communication, nothing prevents the Store from existing on a completely separate process or network node. The Store hosts all the intermediate data created in the process of remote execution. Networked Pointers require the concept of a Distributed Garbage Collection (DGC) mechanism, which is responsible for tracking and freeing objects from the Store when certain conditions are met.

\textbf{Abstract Syntax Tree - }
The Abstract Syntax Tree (AST) is responsible for resolving the correct remote procedure call paths. The AST is a tree that maps any Python module or object to its desired remote behaviour, handling chained pointer generation and message creation on pointer method invocation. This one to one mirror with the existing Python module system makes it trivial to support nearly any third-party Python library with Syft.

A node in the AST can represent anything that a normal Python package can provide. Each node contains references to the actual module, class or callable object and, at the leaves of this tree, all attributes have a return type which determines the synthesised pointer class that is generated when they are invoked remotely. When the caller invokes a remote node via either a handle to the remote systems AST tree or an existing network pointer, a new network pointer is immediately created of the expected return type. A message is then dispatched to the remote system where the AST is used to locate the original reference and execute it; placing the real result into the Store. Under this model only functionality within Python which is explicitly allowlisted and loaded during startup or later via an explicit loading command, can be executed remotely.

\textbf{Communication Protocol - }
Under the Duet architecture, Syft’s network code only supports peer-to-peer connections between two nodes. To connect two peers, by default Duet initialises an outgoing connection to a STUN server operated by OpenMined. STUN is a technology commonly used by video conferencing software which is designed to allow applications to traverse NAT and firewalls. It works by establishing an outgoing connection first to allow any subsequent external traffic on the same port to be routed to the application which made the outgoing request. The STUN server brokers both connections by allowing each side to establish an outgoing connection first. Once paired, the individual peers establish a connection to each other using WebRTC. This connection uses the DataChannel API from WebRTC over UDP, using DTLS as an integrity mechanism. After connection no further traffic is sent to the STUN server. Additionally the source code and instructions to run a copy of this service is available, and connection to a self run STUN service only requires adding a single URL parameter to the Duet initialisation function.

\textbf{Pointers - }
Pointers are dynamic objects created by the AST to provide proxy control of remote objects. Pointers are usually generated from the known return type of a computation. Pointers map the functionality of a remote object to a local object by wrapping and handling the communication layer. Pointers are also responsible for garbage collecting objects which are no longer reachable. The Data Scientist can create remote objects on the Data Owner's side through messages that are sent to the DO. The implementation of the garbage collection (GC) process relies heavily on the underlying Python GC implementation. When the local pointer, that resides on the DS side, goes out of scope, a special Protobuf message is sent to the store. The sole purpose of this message is to remove the actual object that the DS pointer was referring to. This mechanism assumes that an object created by a Data Scientist would not be referred by another user, but it could easily be extended to use a reference counting mechanism where multiple nodes can hold a pointer to the same object.

\subsection{Libraries}
Syft becomes truly powerful when remote computation is extended with the functionality provided by other high performance computing libraries or privacy enhancing frameworks. These tools can be aggregated to support the core components of structurally transparent information flows.

\textbf{PyTorch -} is our initial support of high performance computing on tensors, and the majority of the PyTorch API can be used in Syft. An important note is that our current architecture is not dependent on PyTorch, and our roadmap includes support for other tensor libraries like Numpy, Tensorflow, JAX and scikit-learn.

\textbf{Opacus - } enables PyTorch models to be trained with respect to privacy with minimal changes in the codebase and infrastructure. It can be used to train PyTorch models (\cite{Opacus}).

\textbf{\citet{tenseal} - } a library that enables machine learning frameworks to work on encrypted data, using homomorphic encryption schemes implemented in \citet{sealcrypto}.

\textbf{SyMPC - } SyMPC is a relatively young secure multi party computation library developed in-house. It can not be used as a standalone piece of software since it highly relies on the communication primitives that can be found in Syft. Because of this, you would need to install SyMPC alongside Syft if you want to use any of the implemented functionalities. 
For the moment, since SyMPC it is still at the beginning of the journey, it has some basic arithmetic operations between secretly shared values and secret/public values. For computing the correct result for simple multiplication and matrix multiplication there are employed some triples (presented in \cite{inproceedings}. These primitives are generated by a Trusted Third Party and in our case we presume it is the node that orchestrates the entire computation. A part of the design decisions and implementation details are taken from the Facebook Research Project - CrypTen (\cite{crypten}). SyMPC aims is to offer the possibility to train/run inference on opaque data using a range of different protocols depending on each Data Scientist use case. As an implicit goal, the library should provide a simple way to implement new protocols, regardless of the underlying ML framework that you use.

\textbf{AriesExchanger -} Aries agents facilitate the construction, storage, presentation and verification of credentials which may be used anywhere in the Hyperledger ecosystem. At it's core, the AriesExchanger allows actors in an information flow (Sender or recipient) to verify attributes about the other based on attestations made by trusted authorities, determined by the context. Aries agents facilitate the zero-knowledge presentation of group membership as described by \cite{camenisch2001efficient}. The Aries agent interface is accessed through the AriesExchanger class. From a Syft perspective, AriesExchanger allows Data Owners to only initiate connections with Data Scientists who have credentials to work with their data as described in (\cite{Abramson_2020}). Similarly, Data Scientists can verify whether remote datasets held by Data Owners comply with certain scheme requirements as attested to by the appropriate authority. Governance systems are defined and then implemented through the definition of credential schemes on a distributed ledger. This infrastructure allows for governance systems to be constructed and enforced without the need for a central governing authority. All trust verification is performed peer-to-peer using the Peer DID Communications (PeerDIDComm) protocol (\cite{PyDentity}).  

\textbf{PyDP - } The application of statistical is a Python wrapper for Google's Differential Privacy project. The library provides a set of $\epsilon$-differentially private algorithms. These can be used to produce aggregate statistics over numeric data sets containing private or sensitive information. With PyDP you can control the privacy guarantee and accuracy of your model written in Python. PyDP is being extended to provide DP training of conventional data science algorithms such as bayesian networks and decision trees with scikit-learn. 

\textbf{Syfertext - } is a library which provides secure plaintext pre-processing and secure pipeline deployment for natural language processing in Syft(\cite{Syfertext}).

\textbf{openmined\_psi - } a Private Set Intersection protocol based on ECDH, Bloom Filters, and Golomb Compressed Sets as described in \cite{angelou2020asymmetric}.

\subsection{Tables}
\bgroup
\def\arraystretch{1.3}
\begin{table}[ht]
\centering
\begin{tabular}{ccccccccccccccc}
\hline
\textbf{Name} & \textbf{Date} & \textbf{Org} & \textbf{Base} & \textbf{PSI} & \textbf{VFL} & \textbf{HFL} & \textbf{DP} &\textbf{HE} & \textbf{SMPC} & \textbf{ZkAC} & \textbf{OPC} \\ \hline
PySyft & Jul17 & OpnMnd & TH,+ & 3rd & Y & Y & 3rd & 3rd & Y & 3rd & Y  \\
TFF & Sep18 & Google & Any & N & N & Y & 3rd & N & 3rd & N & N  \\
FATE & Sep18 & WeBank & TH,+ & N & N & Y & 3rd & N & Y & N & N  \\
LEAF & Dec18 & CMU & TF & N & N & Y & 3rd & N & N & N & N \\
eggroll & Jul19 & WeBank & TF,+ & N & N & Y & 3rd & N & N & N & N \\
PaddleFL & Sep19 & Baidu & PD & Y & Y & Y & Y & N & Y & N & N \\ 
FLSim & Nov19 & iQua & TH & N & N & Y & N & N & N & N & N \\
DT-FL & Nov19 & BIL Lab & TH & N & N & Y & N & N & N & Y & N \\
Clara & Dec19 & NVIDIA & TF & N & N & Y & Y & N & N & N & N \\
DP\&FL & Feb20 & SherpaAI & TF,SKL & N & N & Y & Y & N & N  & N & N \\
IBMFL & Jun20 & IBM & KS+ & N & N & Y & Y & N & N & N & N \\
FLeet & Jun20 & EPFL & TF? & N & ? & ? & Y & N & N & N & N \\
IFed & Jun20 & WuhanU & CS & N & ? & ? & Y & N & N & N & N \\
FedML & Jul20& FedML & TH & Y & Y & N & N & N & N & N & N \\
Flower & Jul20 & Cmbrdge & Any & N & N & N & Y & N & N & N & N\\

\end{tabular}
\caption{ Federated Learning Systems Feature Support Comparison. We extend the work of \cite{kritika} to compare PET feature support in existing federated learning architectures. Org: Organisation associated with tool. Base: core machine learning technology. PSI: does the framework provide an api for PSI computation. VFL: does the framework provide an api for some form of Vertically Federated Machine Learning. HFL: does the framework provide an api for some form of Horizontally Federated Machine Learning. DP: does the framework provide an api for some form of Differential Privacy. HE: does the framework provide an api for some form of Homomorphic Encryption. SMPC: does the framework provide an api for some form of Secure Multiparty Computation. ZkAC: does the framework provide an api for some form of Zero-knowledge Access Control. OPC: Does the framework provide an object level RPC. Refs: TFF - \cite{TFF}), FATE - \cite{FATE}, LEAF - \cite{DBLP:journals/corr/abs-1812-01097}), eggroll - \cite{eggroll}, PaddleFL - \cite{PaddleFL}, FLSim - \cite{9155494},  DT-FL - \cite{Abramson_2020}. Clara - \cite{Clara}, DP\%FL - \cite{Sherpa}, IBMFL - \cite{ludwig2020ibm}), DT-FL - \cite{Abramson_2020}, FLeet - \cite{Damaskinos_2020}, IFed - \cite{ifed}, FedML - \cite{fedml}, Flower - \cite{flower}}
\label{table:feature_comparison}
\end{table}
\egroup

\bgroup
\def\arraystretch{1.3}
\begin{table}[ht]
\centering
\begin{tabular}{ccccc}
\hline
\textbf{Forward Step} & \textbf{Processor} & \textbf{Modulus (bits)} & \textbf{File Size} & \textbf{Time Taken} \\ \hline
Input Data & DO & plaintext & 3.47KB \\
Conv1 & DO & plaintext & 10.97 KB & \\
Conv2 & DO & plaintext & 2.4 KB & 27ms \\
Encrypt Signal & DO & 140 & 269.86 KB  & 11 ms\\
\hline
FC1 & DS & 140 & 205.06 KB  \\
Sq. Activation & DS & 140 & 139.6 KB  \\
FC2 & DS & 140 & 68.81 KB & 4.17s \\
\end{tabular}
\caption{Experiment 1, where the DO only receives the convolutional layers. The split layer at FC2 is represented by the bar. As more layers need to be processed, a higher modulus is used. Tests were performed using 4 cores Intel(R) Core(TM) i7-6600U CPU @ 2.60GHz. Forward Step describes the layer of processing the data has just emerged from. The Processor is the actor currently performing the computation. File Size is the size of the signal as it emerged from this layer of processing. Time taken gives the time that has passed since the last time taken. HE parameters; polynomial degree is 8192, coefficient modulus is 140-bits, there is a security level of 128-bits and the scale is $2^{26}$}.
\label{table:experiment1_results}
\end{table}
\egroup

\bgroup
\def\arraystretch{1.3}
\begin{table}[ht]
\centering
\begin{tabular}{ccccc}
\hline
\textbf{Forward Step} & \textbf{Processor} & \textbf{Modulus (bits)} & \textbf{File Size} & \textbf{Time Taken} \\ \hline
Input Data & DO & plaintext & 3.47KB \\
Conv1 & DO & plaintext & 10.97 KB & \\
Conv2 & DO & plaintext & 2.4 KB \\
FC1 & DO & plaintext & 529 B \\
Sq. Activation & DO & plaintext & 529 B & 1ms \\
Encrypt Signal & DO & 88 & 139.62 KB & 4.6 ms \\
\hline
FC2 & DS & 88 & 68.76 KB & 97ms \\
\end{tabular}
\caption{Experiment 2, where the FC1 layer is also sent to the DO. The split layer is represented by the bar. As less layers need to be processed, a lower modulus is used. Tests were performed using 4 cores Intel(R) Core(TM) i7-6600U CPU @ 2.60GHz. Forward Step describes the layer of processing the data has just emerged from. The Processor is the actor currently performing the computation. File Size is the size of the signal as it emerged from this layer of processing. Time taken gives the time that has passed since the last time taken. CKKS parameters; polynomial degree is 8192, coefficient modulus is 88-bits, there is a security level of 128-bits and the scale is $2^{26}$}. 
\label{table:experiment2_results}
\end{table}
\egroup

\end{document}

%% file: math_commands.tex
\usepackage{amsmath,amsfonts,bm}

\def\eqref#1{equation~\ref{#1}}

\def\1{\bm{1}}

\DeclareMathAlphabet{\mathsfit}{\encodingdefault}{\sfdefault}{m}{sl}
\SetMathAlphabet{\mathsfit}{bold}{\encodingdefault}{\sfdefault}{bx}{n}

